# An explainable deep vision system for animal classification and detection in trail-camera images with automatic post-deployment retraining

Golnaz Moallem[a, 1], Don D. Pathirage[a], Joel Reznick[a], James Gallagher[b], Hamed Sari-Sarraf[a]

[a] Electrical and Computer Engineering Department, Texas Tech University, Lubbock, TX, USA 79409

[b] Texas Parks and Wildlife Department, Mason, TX 76856

## Abstract

This paper introduces an automated vision system for animal detection in trail-camera images taken from a field under the administration of the Texas Parks and Wildlife Department. As traditional wildlife counting techniques are intrusive and labor intensive to conduct, trail-camera imaging is a comparatively non-intrusive method for capturing wildlife activity. However, given the large volume of images produced from trail-cameras, manual analysis of the images remains time-consuming and inefficient. We implemented a two-stage deep convolutional neural network pipeline to find animal-containing images in the first stage and then process these images to detect birds in the second stage. The animal classification system classifies animal images with overall 93% sensitivity and 96% specificity. The bird detection system achieves better than 93% sensitivity, 92% specificity, and 68% average Intersection-over-Union rate. The entire pipeline processes an image in less than 0.5 seconds as opposed to an average 30 seconds for a human labeler. We also addressed post-deployment issues related to data drift for the animal classification system as image features vary with seasonal changes. This system utilizes an automatic retraining algorithm to detect data drift and update the system. We introduce a novel technique for detecting drifted images and triggering the retraining procedure. Two statistical experiments are also presented to explain the prediction behavior of the animal classification system. These experiments investigate the cues that steers the system towards a particular decision. Statistical hypothesis testing demonstrates that the presence of an animal in the input image significantly contributes to the system's decisions.

---

[1] : Corresponding author.
Email addresses: golnaz.moallem@ttu.edu (G. Moallem), D.Pathirage@ttu.edu (D. Pathirage), joel.reznick@ttu.edu (J. Reznick), james.gallagher@tpwd.texas.gov (J. F. Gallagher), hamed.sari-sarraf@ttu.edu (H. Sari-Sarraf)

# 1. Introduction

Trail-camera imaging is a non-intrusive method employed in ecological research and conservation to gather large-scale data about wildlife and habitat health [1]. However, the task of manually extracting information from this data is costly, labor intensive, and time-consuming. Moreover, without robust domain expertise, the validity of the produced data is uncertain [2]. Deep neural networks (DNNs) are currently viewed as the state-of-the-art for many computer vision tasks, having made great strides due to advances in computer-hardware, network architectures, and the availability of very large datasets to learn from.

In this work, we propose a two-stage deep learning pipeline for the analysis of wildlife imagery in the Texas Parks and Wildlife Department (TPWD) dataset. In the first stage, a DNN classifies the TPWD images into 'Animal' and 'No-Animal' categories. Then, a second DNN detects and localizes birds in the set of 'Animal' images. Furthermore, this system is managed by an automatic retraining algorithm which maintains performance as data drifts over time. We also present statistical experiments to address model explainability, i.e., insights into network predictions and behavior.

The paper makes the following contributions:

1. It uses off-the-shelf techniques to successfully solve the animal classification and detection problems, which are shown to be unsolvable for our dataset by existing strategies.
2. It uses novel methods for detecting and coping with data drift under realistic field conditions.
3. It employs hypothesis testing to address the explainability of the devised deep network.

The two-stage approach efficiently processes large amounts of data by first filtering out No-Animal images prior to the bird detection phase. This is advantageous as the DNN classifier performs noticeably faster than the DNN detector (details are reported in the following sections); feeding only the Animal-labeled images to the DNN detector reduces the overall analysis time considerably.

Section 2 describes the TPWD dataset. Section 3 elaborates on the training and performance of the classification DNN, i.e. animal classification system. Section 4 introduces an automatic procedure designed for the automatic retraining of the animal classification system. Section 5 presents two statistical experiments explaining the predictions of the animal classification system. The training process and performance of the detection DNN, i.e. bird localization systems, is demonstrated in Section 6. Sections 7 and 8 respectively present discussion and conclusion of the study.

# 2. Dataset

The TPWD dataset is derived from a project investigating the use of prescribed fire to manage wildlife habitat at small scales. While the Northern Bobwhite Quail was the focal species, it was also important to document changes in habitat use by other species of wildlife, with a particular focus on other species of birds.

Traditional wildlife-count techniques would have been difficult to conduct on numerous locations; therefore, trail cameras were used to study wildlife activities at several sites with solar powered water fountains that attract wildlife to the trail camera focal area. Cameras were set to be a standard distance above the fountain (1.52 m) with the same distance from the camera to the fountain (3.05 m). The first year of the study (2014) generated approximately 700,000 images. These images were manually classified by one individual over the course of about 9 months. Given the large size of the image dataset, the large rate of incoming input images, and the need for recurrent image classification, it was necessary to automate this process with high sensitivity and accuracy levels.

This research developed deep neural network (DNN) models for detecting animals, especially birds, in TPWD trail-camera imagery. A significant portion of the work undertaken in this endeavor went into generating useful training and testing datasets from the images provided by TPWD. The images were produced from a set of observation sites which resemble each other in their layout. At each location, a motion-triggered camera placed above the ground was centered and focused on a fountain (artificial watering hole) that attracts animals in the vicinity. Over the course of seven days, the camera continuously monitored the scene for activity, taking images when motion was detected and occasionally, at periodic intervals for diagnostic purposes. Images were recorded for 7-day periods in May, June, and July of each year. For night-time imaging, an infrared (IR) flash was used to illuminate the scene without disturbing the animals. The night-time images are captured by an IR sensitive detector on the camera. Typical examples of night and day images are shown in Figure 1.

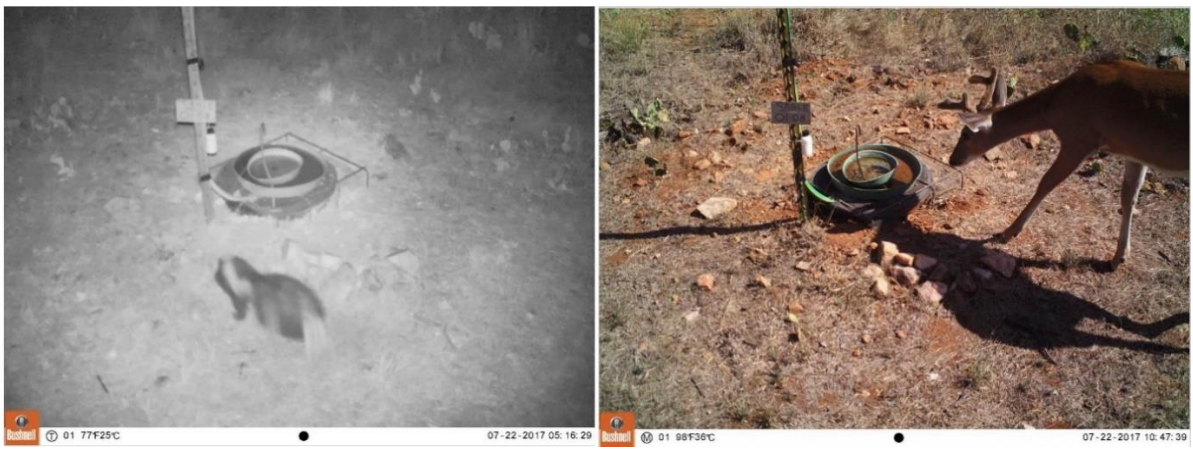

*Figure 1: Illumination from the IR flash in night images (left) washes out much of the background content and in the absence of animal activity, produces a scene that is visually consistent from image to image. The day images (right) vary significantly in background content with changing weather patterns and time of day.*

Given the significant difference in appearance between day and night images, we trained two separate DNNs to analyze the two sets of images. The day-time and night-time datasets used in training and testing these networks were formed from a validated subset of 23,429 volunteer-labeled images, of which only 1,582 contained animals. Figure 2 shows an example of an annotated image in which the animals are labeled and localized with bounding boxes.

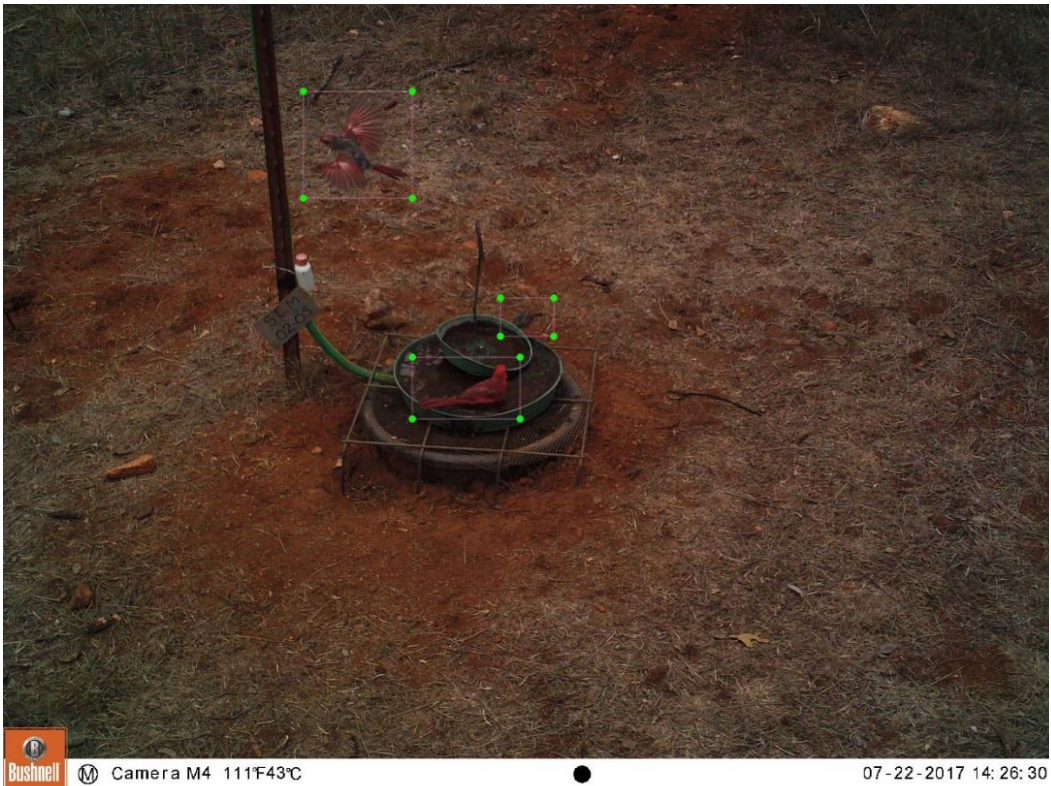
*Figure 2: Images containing animals were annotated with bounding boxes and associated with labels denoting the class the animal in each bounding box belongs to, e.g., mammal, bird, reptile, etc. Images without animals were marked as 'empty'.*

## 3. Animal Classification System

Several other works have employed DNN models for classifying wildlife images from camera-trap projects. We began our work by assessing the results of two such papers by Norouzzadeh et al. [3] and Tabak et al. [4] which outline methods for classification of larger mammals (compared to those in the TPWD images) in images from the SnapShot Serengeti (SS) project [5]. We applied the DNN models produced from these works to analyze images from the TPWD dataset. Despite the shared domain relevance between the datasets and similar classification tasks, the networks performed poorly on a benchmark set of TPWD images — in that nearly all images containing no animals produced false positive predictions.

To expedite the development of a more performant DNN, we also explored methods of leveraging transfer learning from a larger, already annotated dataset having better domain overlap with the TPWD images. For this, we trained models on images from the iWildCam 2018 challenge dataset [6] which tracks animals and geographies that are more comparable to those observed in the TPWD images. Like the SS networks, these models too generated mainly false positive predictions. Observing this pattern, we speculated that the presence of the watering fountain, common to all the TPWD images, may be triggering false positive detections. To verify this, we applied inpainting with Nvidia's Inpainting DNN [7] (see Figure 3) to remove the watering fountain from images with no animals and observed that the networks began to classify such images as true negatives.

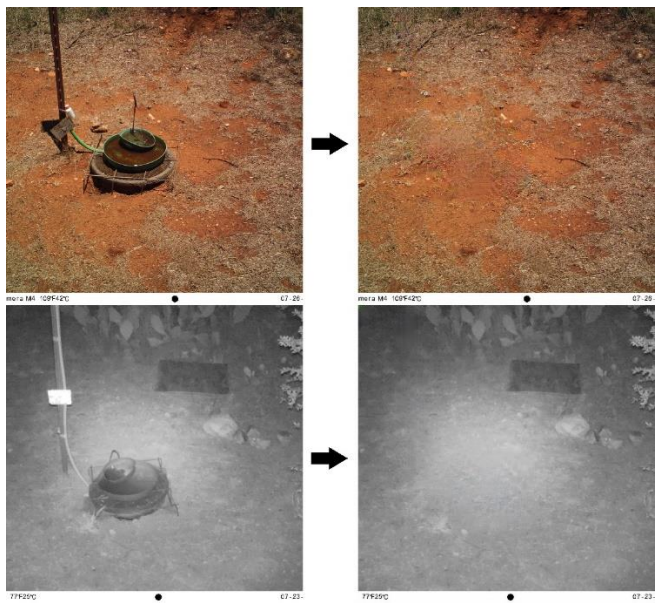

*Figure 3: Watering fountain in the images was masked out using Nvidia's Inpainting DNN [7].*

Given the apparent bias with existing DNNs toward background scene information, even in cases with significant domain overlap, it was evident that new models needed to be trained specifically on the TPWD images. This necessitated the laborious endeavor of annotating the TPWD images to generate training and testing datasets.

### 3.1. Dataset Generation Procedure

Initial experiments with random sampling of the labeled TPWD images to generate balanced datasets of Animal and No-Animal classes resulted in models that were highly sensitive to the background content and day-time shadow patterns, which occur naturally in the scene. Consequently, the developed models were again producing mainly false positive detections. We therefore aimed to develop models which better accounted for the variation in background content and shadow patterns by applying a more appropriate procedure for generating training data.

Furthermore, due to the severe imbalance between Animal and No-Animal examples in the TPWD images (1,662 Animal and 21,847 No-Animal), special emphasis was also placed on ensuring the sampling procedure produced balanced and representative modeling datasets to prevent the DNNs from becoming biased towards background information or a particular class.

Data augmentation has been shown to play a critical role in producing effective predictive models for visual tasks but requires domain-specific knowledge on when and how to apply the augmentation techniques [8]. To amplify the number of animal examples in the training data and define a robust predictive task, Animal images were augmented by flipping horizontally about the central y-axis; see Figure 4. The augmented dataset aims to produce models that are invariant to whether an animal appears in a left or a right profile in the image. We considered other augmentations such as rotations, additive noise, and blurring but did not find them to be as useful as horizontal image flipping. After augmenting the Animal images, a roughly equal number of time-sampled background images displaying shadow patterns from each observation site was incorporated into the training dataset for the No-Animal class.

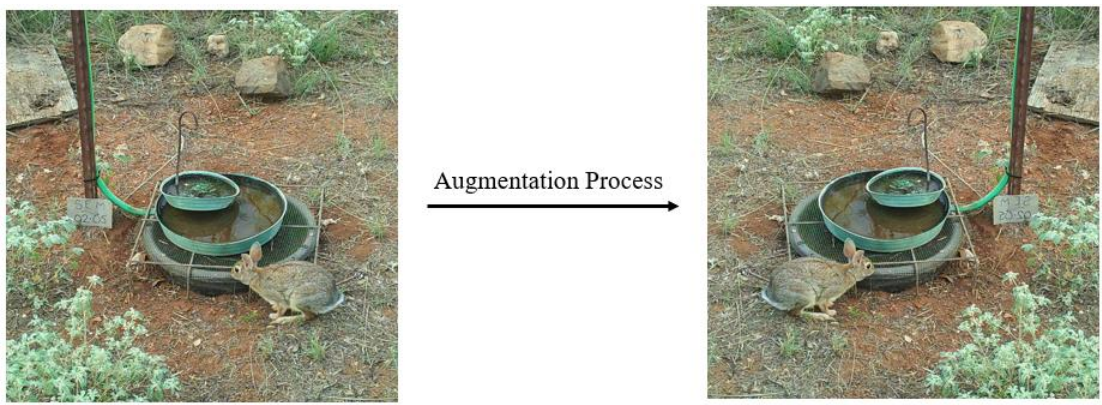

*Figure 4: Animal images in the training set were augmented via horizontal flipping about the central y-axis.*

As our models expect an input image size of 299 × 299 pixels, the original 3264 × 2448 images needed to be resized, but simply resizing these large images can lead to problems, e.g., pixels of very small animals such as birds (the majority of animal examples in day images) will be decimated or lost after resizing. To address this problem, day-time images for both training and testing were first cropped using a 1500 × 1500 window centered over the watering fountain — the region of the image where animal activity is highest as determined from the frequency of bounding box labels. We selected the size and region of this cropping window using two criteria: (1) The size of the window after resizing to 299 × 299 should not negatively impact the accurate classification of the smallest animals, i.e., birds, and (2) the cropped dataset should retain at least 90% of the original animal examples. Our choice of 1500 × 1500 window centered around the watering fountain resulted in an acceptable 9% loss of day-time animal examples from the original, uncropped dataset. The window cropping algorithm is shown in Figure 5. In contrast, as there was little to no bird activity in the night-time images, the same procedure was not necessary for the night-time training dataset.

An additional criterion was used in selecting No-Animal images for the day-time dataset — they needed to be well-representative of the various lighting conditions and shadow patterns that occur at each location. This was accomplished by employing a time-of-capture based sampling of images for each location in the dataset. Animal examples were sampled from a histogram with 15-minute interval bins and the No-Animal examples from a histogram with 3-minute interval bins. Time-of-capture sampling was not used for the night-time datasets as there was minimal variation in the background due to the very consistent illumination provided by the IR flash.

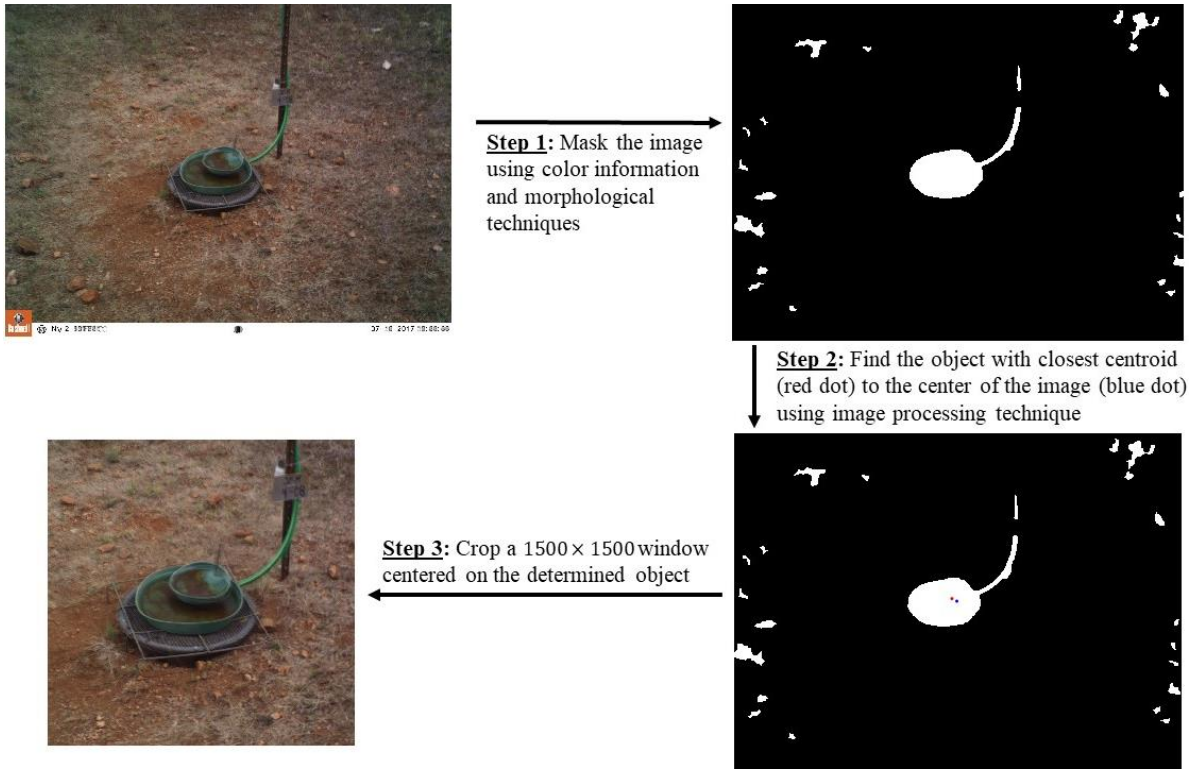

*Figure 5: The automatic window cropping algorithm*

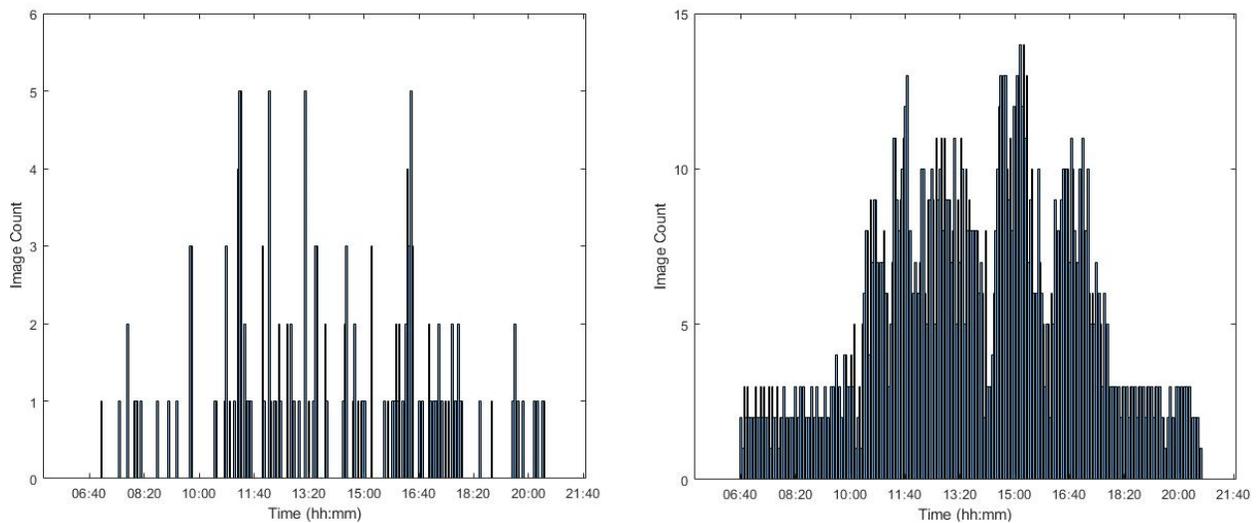

*Figure 6: Histograms show, respectively, the typical distributions of Animal (left) and No-Animal (right) images from a single location. For the No-Animal class, we used the images captured at set intervals (for diagnostic purposes) to select a set of background images that are well representative of the shadow patterns that occur at each location.*

### 3.2. Transfer Learning and Architecture Selection

To speed up the development of models specific to our task and dataset, we applied concepts from transfer learning to existing state-of-the-art network architectures. In contrast to other works mentioned in this paper, we found this step necessary as our datasets were not large enough to train models from the ground up. The architecture for our models was selected by comparing pre-trained network performances on ImageNet [9], a large object classification dataset consisting of over 14 million images for more than 20,000 classes. The assumption is that the convolutional layers of a pre-trained, well-performing network on ImageNet will also be suitable for datasets in our domain as these layers learn features such as edges and textures that are common in all object detection tasks. The size and parameter count of these networks' layers also influenced architecture selection as computational resources required to re-train a model were limited. The main limiting factor here was the amount of VRAM available in our 11GB GPU to handle the volume of data necessary for re-training our models. Following these considerations, a pre-trained Xception architecture [10] was selected. The Xception architecture achieves greater than 90% Top-5 Accuracy on ImageNet with lower model and computational complexity than other networks, making it a good candidate for modeling our classification task [11]. Training our day-time and night-time Xception models on an Nvidia GTX 1080 Ti took 50 mins and 42 mins, respectively. Inference time for both models was 5.5 ms on an AMD Ryzen 3900X.

### 3.3. Animal Classification Results

We trained two separate Xception networks, one for day-time images and one for night-time images. After determining whether an image is from day-time (high hue value) or night-time (near-zero hue value) by comparing them in the HSL (hue, saturation, lightness) color-space, our models take the input image and output two probabilities for whether the image contains an animal or not.

Of the 8 observation sites in our dataset, 5 were discarded from use in training the day-time model because they do not have enough animal examples to generate datasets that are both balanced and well-representative of each site's background content and shadow patterns. We trained and tested the day-time network on 3,085 images from the remaining three observational sites. Conversely, the background is generally uniform from site to site in the night images, so the night-time network was trained and tested on 1,036 images from all 8 observation sites. Note that the inclusion of any number of sites greater than one in training the two models is sufficient to demonstrate the ability to build predictive models that are site-invariant. The generated training sets are outlined in Table 1. Our best models achieved 94% and 98% classification accuracy on the benchmark dataset for day-time and night-time images, respectively. Day-time predictions had a sensitivity of 87% and a specificity of 96%. Night-time predictions had 99% and 96% sensitivity and specificity, respectively. Details of each model's performance and their testing sets are presented in Table 2.

Additionally, we combined the day-time and night-time training datasets to train a single classification model using the same Xception architecture. The combined model performed with an overall 91% accuracy on the benchmark dataset whereas, the individual day-time and night-time models had an overall performance of 96%. Given this performance delta, we opted to use the two separate model approach for day-time and night-time images in the ACS.

Using the time-of-capture sampling strategy to incorporate the variation in shadow patterns and background content of observation sites into the day-time training data alleviated the problem of frequent false positive detections caused by training on only randomly sampled data. In contrast, the visual uniformity within and between observation sites from the night-time images made the task of training a classification network simpler.

Table 1: Statistics of the day-time and night-time training sets

| Obs. Site | Day-time Training Set | | | |
|---|---|---|---|---|
| | Site No. 1 | Site No. 2 | Site No. 3 | Total |
| 'Animal' class | 288 | 268 | 393 | **949** |
| 'No-Animal' class | 281 | 289 | 327 | **897** |

| Obs. Site | Night-time Training Set | | | | | | | | |
|---|---|---|---|---|---|---|---|---|---|
| | Site No. 1 | Site No. 2 | Site No. 3 | Site No. 4 | Site No. 5 | Site No. 6 | Site No. 7 | Site No. 8 | Total |
| 'Animal' class | 46 | 47 | 91 | 21 | 89 | 18 | 12 | 44 | **368** |
| 'No-Animal' class | 46 | 47 | 91 | 21 | 89 | 18 | 12 | 44 | **368** |

Table 2: Statistical measures of the models' performance are presented in the form of Sensitivity and Specificity for each observation site. TP, TN, FP, and FN refer to true positive, true negative, false positive, and false negative, respectively.

| Daytime Model | | | | | | | | | |
|---|---|---|---|---|---|---|---|---|---|
| 'Animal' Class | | | | | 'No-Animal' Class | | | | |
| Obs. Site | Site No. 1 | Site No.2 | Site No. 3 | Total | Obs. Site | Site No. 1 | Site No.2 | Site No. 3 | Total |
| # of Images | 110 | 113 | 175 | **398** | # of Images | 268 | 262 | 311 | **841** |
| TPs | 97 | 102 | 147 | **346** | TNs | 261 | 254 | 294 | **809** |
| FNs | 13 | 11 | 28 | **52** | FPs | 7 | 8 | 17 | **32** |
| Sensitivity | **88%** | **90%** | **84%** | **87%** | Specificity | **97%** | **97%** | **95%** | **96%** |

| Nighttime Model | | | | | | | | |
|---|---|---|---|---|---|---|---|---|
| Obs. Site | 'Animal' Class | | | 'No-Animal' class | | | Sensitivity | Specificity |
| | # of Images | TP | FN | # of Images | TN | FP | | |
| Site No. 1 | 20 | 20 | 0 | 20 | 17 | 3 | 100% | 85% |
| Site No. 2 | 19 | 18 | 1 | 20 | 20 | 0 | 95% | 100% |
| Site No. 3 | 38 | 38 | 0 | 39 | 39 | 0 | 100% | 100% |
| Site No. 4 | 9 | 9 | 0 | 9 | 9 | 0 | 100% | 100% |
| Site No. 5 | 34 | 34 | 0 | 34 | 32 | 2 | 100% | 94% |
| Site No. 6 | 7 | 7 | 0 | 7 | 6 | 1 | 100% | 86% |
| Site No. 7 | 5 | 5 | 0 | 5 | 5 | 0 | 100% | 100% |
| Site No. 8 | 18 | 18 | 0 | 16 | 16 | 0 | 100% | 100% |
| **Overall** | **150** | **149** | **1** | **150** | **144** | **6** | **99%** | **96%** |

# 4. Automatic Retraining Procedure

A crucial characteristic of a reliable and robust deep learning system is its ability to generalize and respond in a stable fashion to drift in the incoming data. Ideally, once deployed, a robust system continuously monitors the incoming data and detects any drift in the data that may lead to performance degradation and if necessary, triggers a retraining procedure.

In the TPWD images, the observation sites can look noticeably different with the passage of time and changes in environmental conditions (Figure 7). In these images, data drift manifests itself as background changes in the observation sites. More specifically, the drifted images contain components that the model (1) is not trained for and (2) can switch the model's classification decision. Such components hereafter will be referred to as notable background changes. These components are mainly temporally and statistically dependent background objects that are added, eliminated, or have appearance or location transformation unseen to the model. Examples of such components may be change in the background vegetation state, displacement of existing objects (e.g., big rocks, cardboards, buckets, or watering fountain), and the introduction of new objects into the scene. All such factors can potentially transform the background scene in a significant way and cause the deployed model to produce false positives.

To assess the post-deployment health of the Animal Classification System (ACS), we first trained it on a subset of TPWD images from July 2017 following the procedures outlined in section 3. We then tested this system, referred to hereafter as ACS 2017, on a set of randomly selected images from 2019.

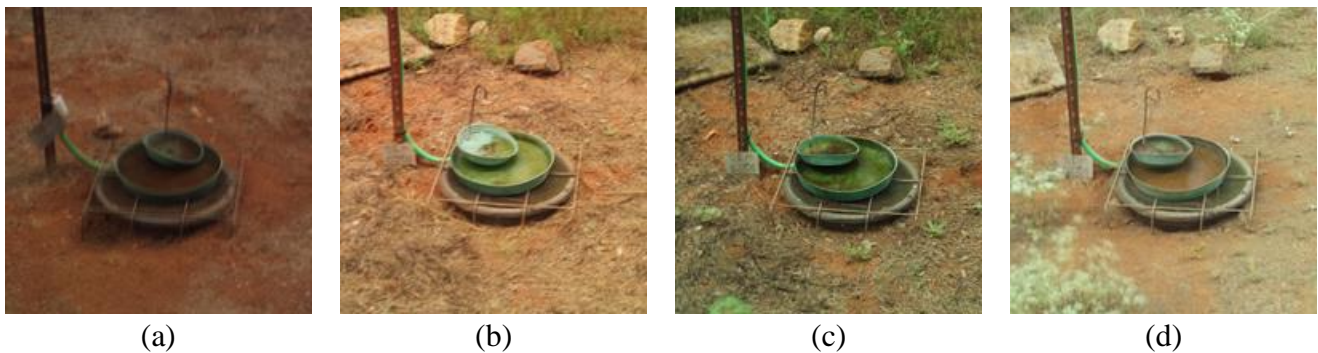

|   (a)   |   (b)   |   (c)   |   (d)   |

*Figure 7: One of the observation sites (site No. 3) in chronological order: (a) July 2017, (b) May 2019, (c) June 2019, and (d) July 2019*

The deteriorated performance of ACS 2017, reported in Table 3, indicates that incoming images gradually drift as the appearance of the background changes over time. Furthermore, degradation is more pronounced in day images where background content plays a prominent role than in night images where the background is generally uniform in appearance. Given these results and our assumptions about the effects of background changes on performance, we developed a technique to detect and quantify such changes.

Table 3: Post-deployment performance of ACS 2017 on the 2019 test set. Deterioration of performance of this model as compared to that reported in Section III is reported in parenthesis.

| Day-time Model | | | Night-time Model | | |
|---|---|---|---|---|---|
| Sensitivity | Specificity | Y-index | Sensitivity | Specificity | Y-index |
| 81.4% (-5.4%) | 77.7% (-19.8%) | 61% (-23%) | 94.8% (-4.2%) | 95.8% (-0.1%) | 91% (-4%) |

A deployed model is capable of handling backgrounds and components that are adequately represented in the training images. Drifted images are those with notable background changes, i.e., backgrounds or content that deviate significantly in appearance, compared to the training images. We quantify these notable background changes by comparing background states of incoming images against the background states in the training data to determine whether performance may be impacted.

However, temporally- and statistically-independent components in the background such as animal presence, shadow patterns and vegetation movements also affect the background comparison, even for images containing very similar background states. Consequently, a one-to-one comparison of individual incoming images and training images is not practical. To resolve this problem, mean images were introduced which essentially eliminate the temporally- and statistically- independent components present in image to image. Mean images of observation sites for a specific time interval were calculated by averaging all the cropped images taken from the corresponding observation site during that time interval.

Based on observations from over 10,000 images in our dataset, the background of a site did not go through notable changes from one sunrise to sunset. Therefore, the means of both incoming and training images from sunrise to the sunset within a day for each observation site was estimated and used for comparison.

The goal was to determine if the model is trained for the background state in the incoming images. If not, trigger the retraining process. Triggering of automatic retraining was accomplished through the following steps per observation site:

1. For a day worth of incoming images, estimate the mean image ($\bar{I}_{one\_day}$).

2. Assuming there are $N$ background states available in the training set of the deployed model, compare $\bar{I}_{one\_day}$ to these training background states ($\bar{I}_{BG\_state\_1}, \ldots, \bar{I}_{BG\_state\_N}$). Each background state is the mean of the training images captured on a single day.

3. If $\bar{I}_{one\_day}$ is **similar** to one of $\bar{I}_{BG\_state\_1}, \ldots, \bar{I}_{BG\_state\_N}$, then that means the model is trained for the background state in that day and theoretically can perform the classification task adequately for those images. Otherwise, it requires retraining.

The measure of similarity between $\bar{I}_{one\_day}$ and background states is quantified as follows; we set up a technique employing the structural component of the similarity index (SSIM) [12] as defined by:

$$s(x,y) = \frac{\sigma_{xy} + C_3}{\sigma_x \sigma_y + C_3}, \quad (1)$$

where $x$ and $y$ are images being compared, $\sigma_{xy}$ is the cross-correlation of $x$ and $y$, $\sigma_x$ and $\sigma_y$ are the standard deviation of $x$ and $y$, respectively, and $C_3$ is the regularization constant. This component contains the structural similarity information defined as the luminance- and contrast-independent characteristics that account for the structure of objects in the field of view [12].

To determine the similarity between two mean images, local SSIM-Structure values were calculated for corresponding sub-regions between the mean images. Because the key background feature, i.e., the watering fountain, occupies a $500 \times 500$ neighborhood in all $1500 \times 1500$ images, local structural calculations were performed within $500 \times 500$ windows with a stride of 250 pixels and the results of this procedure were stored in a $5 \times 5$ SSIM-Structure matrix.

Figure 8 demonstrates two examples of corresponding sub-regions from mean images of observation site No. 3 over different time intervals. The presence of rocks and vegetation in the sub-region from July 2019 shown in Figure 8(b) causes discrepancies in the structure of the scene that result in a relatively low SSIM-Structure value. On the other hand, sub-regions in Figure 8(c) and 8(d) contain similar structural components and, hence, produce a high SSIM-Structure value.

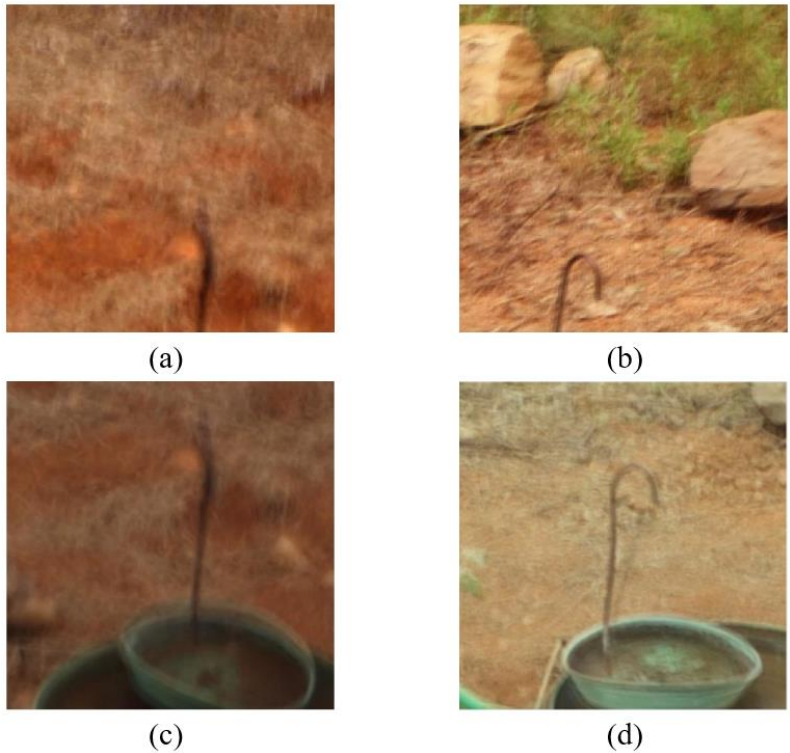

*Figure 8: Examples of corresponding sub-regions from two mean images of observation site No. 3. The sub-regions (a) and (c) are cropped from the July 2017 mean image and the sub-regions (b) and (d) are extracted from July 2019 mean image. The estimated SSIM-Structure for sub-regions (a) and (b) is 0.37, whereas the calculated SSIM-Structure for sub-regions (c) and (d) is 0.84.*

We chose the standard deviation of the SSIM-Structure matrix for measuring the dissimilarity of mean images. This measure is referred to as the Retraining Trigger Index (RTI).

Figure 9 displays the heatmap of estimated RTI values for several pairs of $\bar{I}_{one\_day}$ images from observation site No. 1. As expected, all the diagonal components are zero, because an image is compared to itself and so all the elements of the SSIM-Structure matrix are ones. Therefore, the standard deviation of this matrix defined as the RTI is zero. Moreover, the RTI values associated with intra-monthly pairs are noticeably smaller than those of inter-monthly pairs since in terms of vegetation growth, month-to-month background changes are more drastic compared to background changes that occur within a month. The low RTI value of Pair 1 validates the visual similarity between the two mean images. The higher RTI of Pair 2 compared to Pair 1 shows that Pair 2 manifests more local dissimilarities. However, both these pairs have RTI values less than 0.1 and neither carries a notable background change. On the other hand, Pairs 3 and 4, associated with RTI values above 0.1, exhibit notable background changes, e.g., vegetation state change and displacement of the fountain.

Inspecting 602 mean image pairs visually and monitoring their associated RTIs, we found that RTI values higher than 0.1 indicate a notable background change. Accordingly, the retraining triggering procedure is illustrated in Figure 10. Every time an $\bar{I}_{one\_day}$ triggers retraining, a subset of images associated with that $\bar{I}_{one\_day}$ is formed by sampling from the associated temporal histogram. (see Section 1). This subset is then appended to the model's training set, and the model is retrained with the enhanced training set.

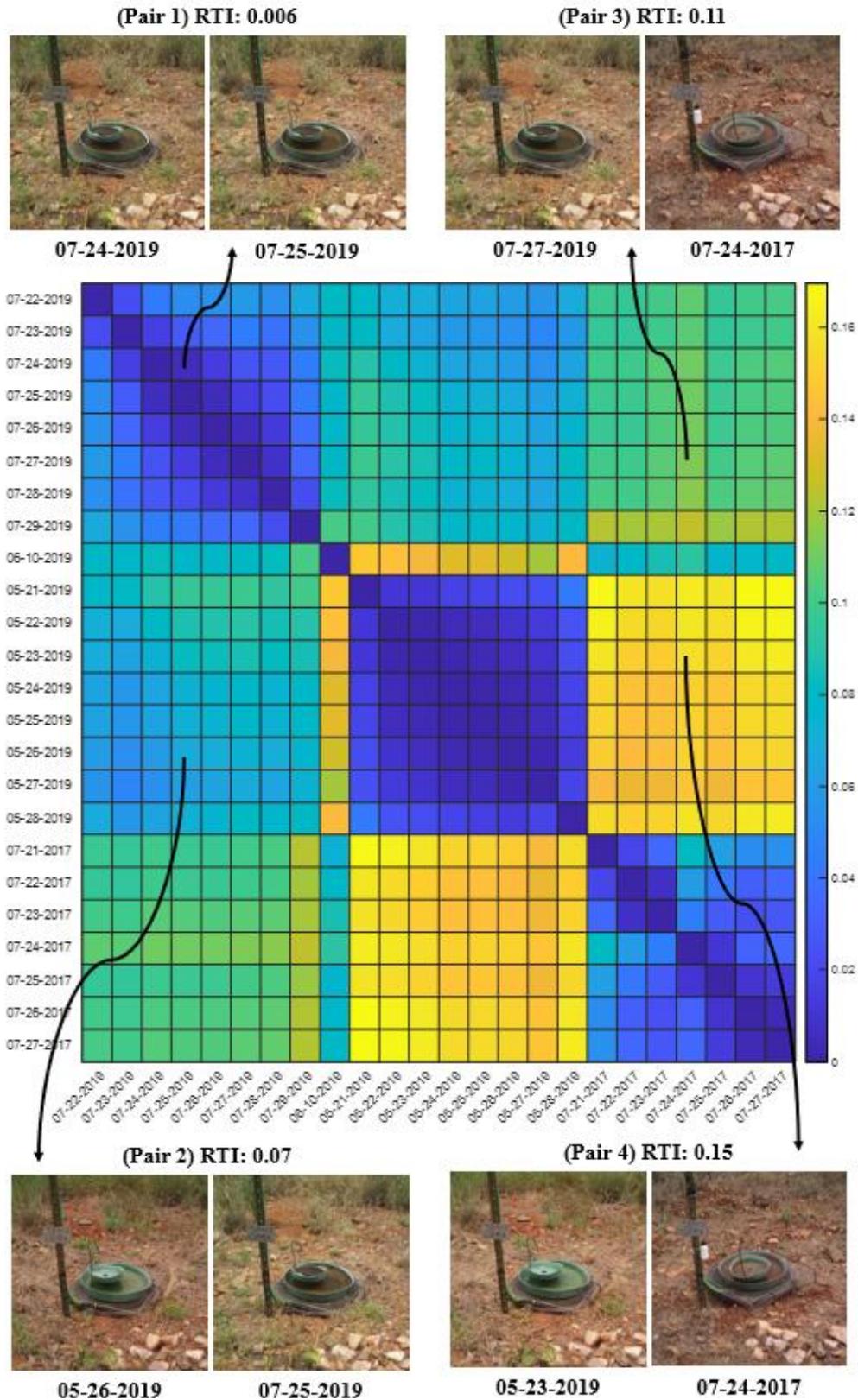

Figure 9: Estimated RTI values for pairs of $\bar{I}_{one\_day}$ images in observation site No. 1. Higher values are shown by lighter color as lower values are illustrated by darker colors.

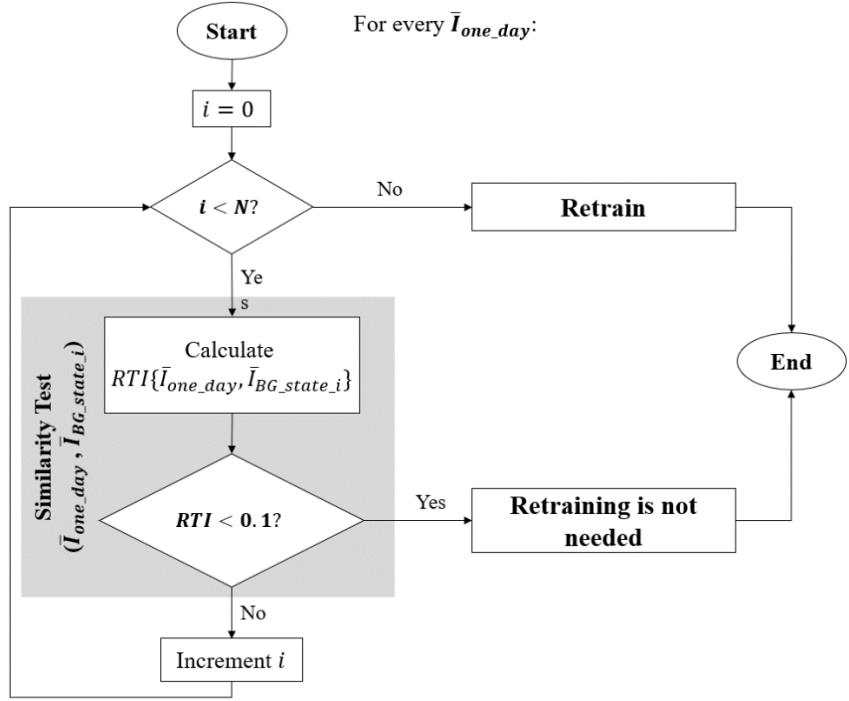

*Figure 10: The algorithmic flowchart of the automatic retraining triggering procedure. For the similarity test of $\bar{I}_{one\_day}$ and $\bar{I}_{BG\_state\_i}$, the RTI value is estimated and thresholded.*

To demonstrate how this retraining triggering system works, we deployed the ACS 2017 model accompanied with eight training background states on two sets of incoming one-day images:

1. Figure 11 illustrates the steps of the retraining triggering algorithm for the $\bar{I}_{one\_day}$ from July 2019. The background state of $\bar{I}_{one\_day}$ is compared with all the training background states ($\bar{I}_{BG\_state\_1}, \dots, \bar{I}_{BG\_state\_8}$) available in ACS 2017 and the RTI values are estimated. One may easily observe that the $\bar{I}_{one\_day}$ has notable background changes through visual inspection; the estimated RTI values are all greater than 0.1. Here, the algorithm recommends that the model needs retraining. To determine if this recommendation is reasonable, we tested the ACS 2017 model on a subset of the images associated with $\bar{I}_{one\_day}$, which resulted in a poor 67% sensitivity and 60% specificity. Following the recommendation, the retrained model achieved a sensitivity and specificity of 100% for the same images, further confirming the algorithm's recommendation.
2. Figure 12 demonstrates the same process for the $\bar{I}_{one\_day}$ from July 2017. The computed RTI for the first background state ($\bar{I}_{BG\_state\_1}$) is 0.03. The RTI value being less than 0.1, the algorithm's recommendation is that retraining is not necessary; The model is already trained to handle the background state of $\bar{I}_{one\_day}$. The model was again tested on the images affiliated to $\bar{I}_{one\_day}$, resulting in 81% sensitivity and 95 % specificity. These results validate the algorithms recommendation.

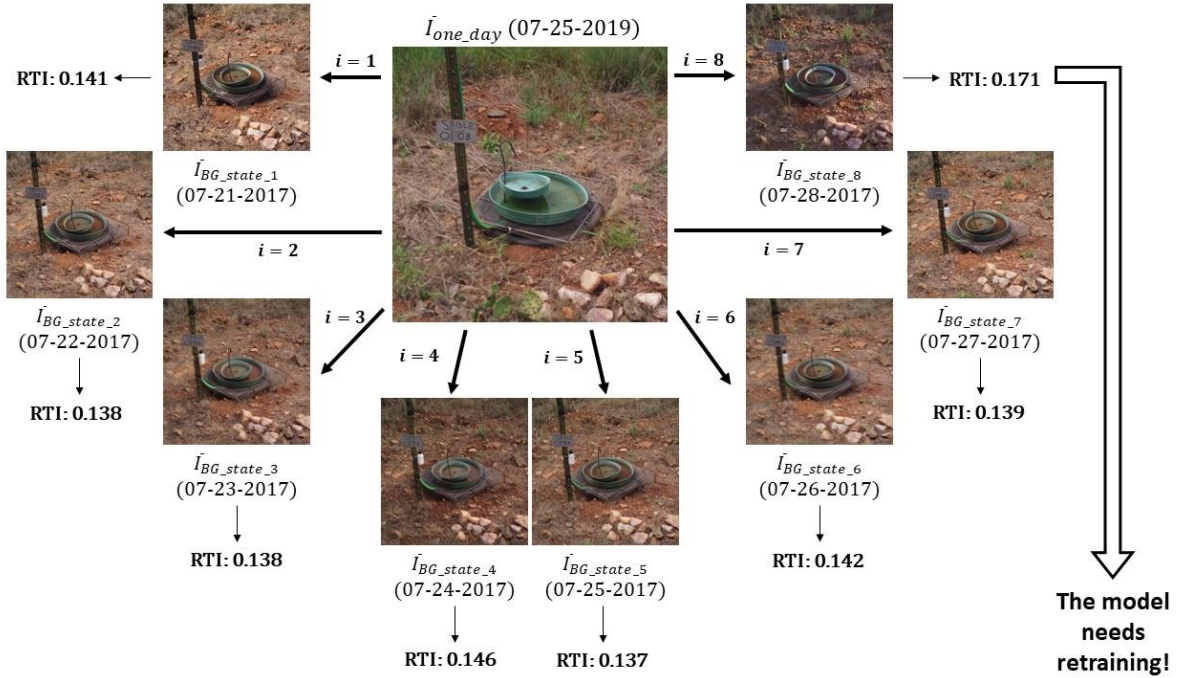

*Figure 11: Retraining triggering procedure on the single-day images captured in July 2019*

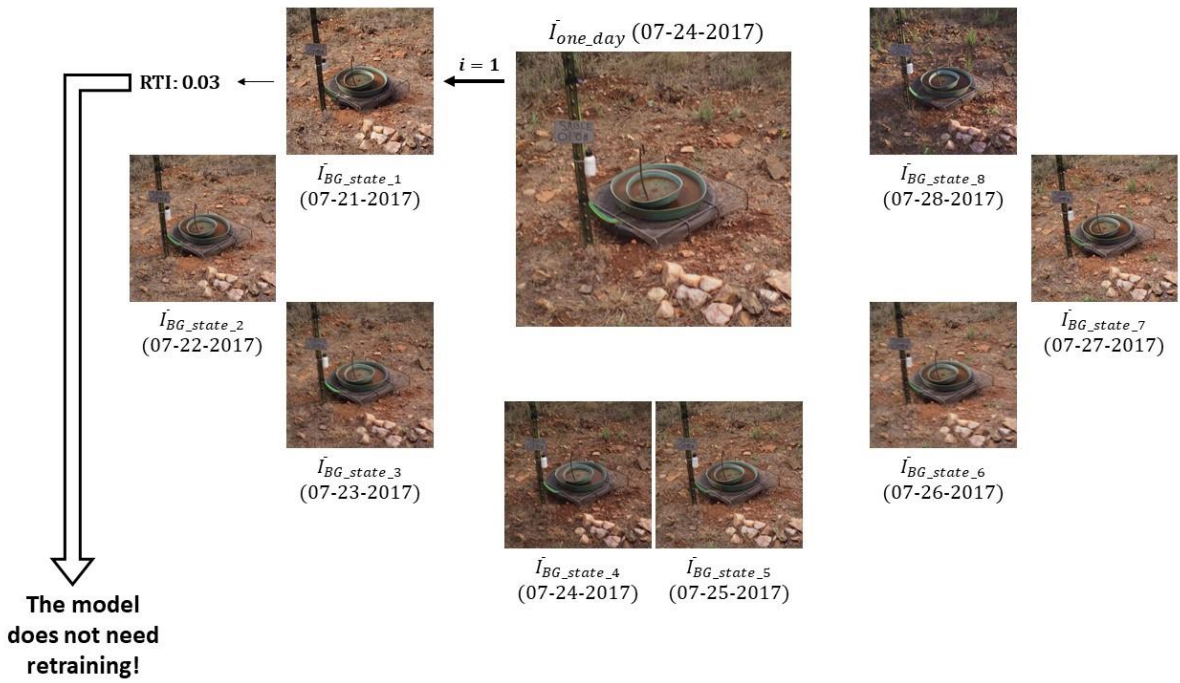

*Figure 12: Retraining triggering procedure for the single-day images taken in July 2017*

Based on these experiments and observations, the automatic retraining triggering algorithm employing the introduced RTI is shown to be a reliable technique for enabling the ACS to adapt to data drift and maintain robust performance.

# 5. Explainability

The CNNs have demonstrated remarkable success with various image classification tasks [13-16]. As shown with the ACS, an *adequately trained* model is very successful at classifying various animal species over several observation sites. However, the way in which the ACS arrives at a particular decision is not readily transparent; specifically, the criteria and features in an input image considered by the CNN models to determine a classification label.

This major shortcoming in the interpretation of a CNN classification system originates from the black-box nature of deep learning networks. This subject has been recently addressed in several works [17-28]. There have been several visualization tools and libraries developed for explaining deep Neural Networks [20, 22, 23]. Moreover, heatmap visualization approaches have been used in explaining the decisions of the deep neural networks [18, 26-28]. While these methods propose a general explanation for how a trained system works; this section introduces a focused interpretation of the CNN classifier in the ACS using a frequentist statistical approach. We propose two statistical experiments to investigate the rationale of the network behind its correct decisions, as follows:

I. True-Positive (TP) experiment, which investigates the motive behind the classifier's decisions for TP images
II. True-Negative (TN) experiment, which examines the rationale behind the classification of TN images

The following presents a detailed description of these experiments. The experiments are executed on $1500 \times 1500$ cropped day-time images from both ACS training and benchmark datasets. This collection of images is referred to as the "experimental set". The reason we did not conduct a similar hypothesis test for the night-time model is two-fold. First, since the day-time and night-time models have identical architectures and since the night-time images are structurally less complex, we believe that employing the same statistical experiment with the night-time model would produce predictably similar results. Second, the night-time testing set is too small for a meaningful statistical analysis.

## 5.1. True-Positive (TP) Experiment

In the True-Positive experiment, the performance of the ACS on TP images is analyzed. TP images are images in the experimental set containing an animal in the field of view that are correctly classified into the 'Animal' category. For this experiment, we posit the following:

Null hypothesis ($H_0$): The ACS <u>significantly</u> bases the classification decision (Animal/No-Animal) on the presence of an animal in the input image.

The alternative hypothesis is, therefore defined as:

Alternative Hypothesis ($H_a$): The ACS bases the classification decision regardless of the presence of an animal in the input image.

The data preparation phase for this experiment is rather cumbersome, yet doable if performed in an organized manner. Table 4 and Table 5 describe such a workflow. Every TP image is paired with a No-Animal image based on the temporal and structural aspects. We simply refer to the paired image as the *twin image*. The algorithm for finding the twin image is demonstrated in Table 5.

*Table 4: The algorithm designed for True-Positive Experiment*

---
**Experiment I: True-Positive Experiment**

**Input:** $N_{tp}$ experimental images correctly labeled as "Animal", i.e., TP images
**for** *TP images $i = 1, 2, ..., N_{tp}$* **do**
    1. Find the twin image using Algorithm I
    2. Feed the twin image to the ACS and collect the estimated label
**end**
**Using** the collected statistics**:**
    **Establish** the t-test stat, $t = \frac{\bar{x} - \mu}{\frac{s}{\sqrt{n}}}$

$\bar{x}$: sample means
$s^2$: sample variance
$n$: sample size
µ: specified population means
$t$: Student-t quantile with n-1 degrees of freedom
$p - value$: corresponding calculated probability, defined as the probability of finding the observed results when the null hypothesis (H$_0$) is true
**Reject** the null hypothesis for a significance level $\alpha = 0.05$ if the calculated p-value is less than $\alpha$.
**Otherwise**, the experiment fails to reject the null hypothesis; this simply means that the data supports the null hypothesis. A significant p-value indicates strong support for the null hypothesis.

---

*Table 5: The algorithm designed for finding the twin image*

---
**Algorithm I: Finding the twin image**

**Input:** one TP image & all the No-Animal images from the same observation site
**Define** dissimilarity index (DISI)

$$DISI = \frac{1}{60\ sec} \times \left(T_{TP\ image}(sec) - T_{No-Animal\ image}(sec)\right) + \left(1 - SIM_{TP\ image,\ No-Animal\ image}\right) \quad (1)$$

**DISI:** dissimilarity index, dimensionless
**T:** Timestamp associated with an image, second
**SIM:** Similarity Index **[12],** dimensionless
**1/60:** conversion factor

**for** *No-Animal images $i = 1, 2, ...$* **do**
    1. Calculate the DISI for the $i^{th}$ No-Animal image and the TP image
    2. Record the DISI for the corresponding No-Animal image
**end**
        **No-Animal image with minimum DISI → twin image**

---

The dissimilarity index (DISI), defined in Equation (1), quantifies the degree of temporal and structural similarity. The DISI value for the TP image and a No-Animal test image consists of two terms: (1) the time stamp difference associated with the two images, this signifies the temporal similarity, and (2) the similarity index of the two images as discussed in detail in [12]. Finding the twin image based solely on temporal similarity is not sufficient because not only temporal features such as shadow patterns contribute to the appearance of the observation site but also, other environmental features, e.g., cloud overcast, rain, wind, etc., the impacts of which can be properly quantified by the similarity index.

Figure 13 demonstrates an example of a TP image and its twin image.

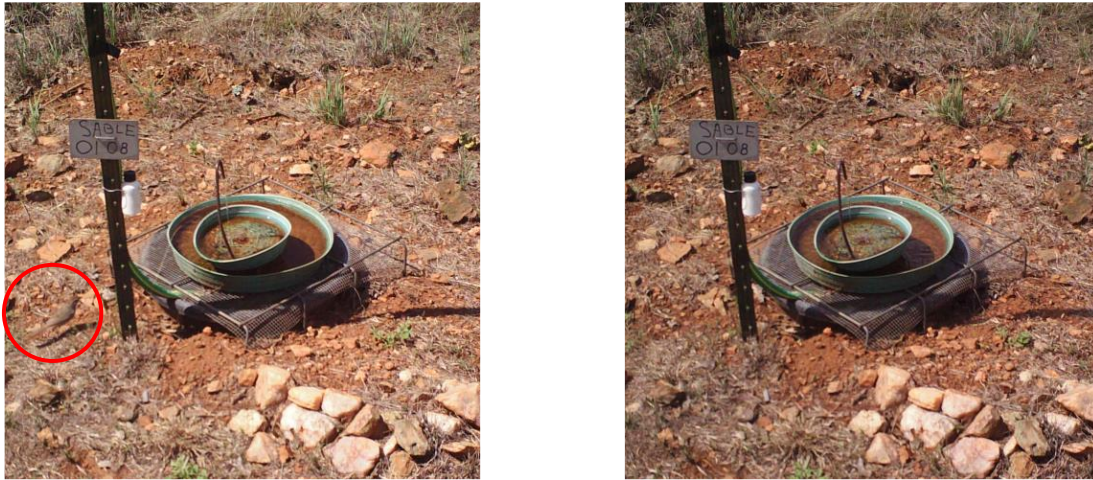

*Figure 13: An example of a TP image (Left) and its twin image (Right). As shown, the two images are almost the same with respect to the background components (watering fountain placement, vegetation, rocks, etc.) and the shadow patterns. The only noticeable difference is the bird in the TP image (circled in red).*

The results of this experiment for all TP images indicate that in at least 94% of the cases, the twin image received a "No-Animal" label (See Table 6). We take the following steps to test the null hypothesis:

1. We assume that an acceptable performance for the ACS on twin images is 0.95. We simply refer to this statistic as the "success rate," denoted as $\boldsymbol{p} = 0.95$. The total number of samples is 1,190.
2. We may think of this process as a series of $\boldsymbol{N_{tp}} = 1{,}190$ binomial samples, for which the expected success rate of correct labeling is 0.95. The samples are assumed to be independent.

An investigation of the underlying binomial distribution justifies a Normal approximation to the binomial distribution. The binomial variance, defined as $N_{tp}p(1-p) = 56.5$ is significantly higher than the threshold level of 10 [29]. This observation justifies a safe application of the normal-based t-test.

*Table 6: The results of the True-Positive experiment. As demonstrated, for all three observation sites, ACS estimates the 'No-Animal' label for the selected twin images at least 93% of the cases.*

| Site | Total twin images | No. of twin images correctly classified | ACS Specificity on twin images |
|---|---|---|---|
| Site No. 1 | 346 | 333 | 96% |
| Site No. 2 | 342 | 322 | 94% |
| Site No. 3 | 502 | 478 | 95% |
| **Total** | **1,190** | **1,133** | **95%** |

A one-sided t-test with a 0.05 significance level confirms a 0.95 minimum success rate of correct labeling. The one-sided t-test fails to reject the null hypothesis with a strong p-value of 0.63 and an upper bound confidence value of 0.962. Therefore, the ACS significantly relies on the presence of an animal to pass an 'Animal' label. Note that, if the expected success rate is dropped to 0.94, the t-test would still fail to reject the null hypothesis with notably stronger p-value of 0.97.

Consequently, it can be confidently concluded that the ACS significantly emphasizes the presence of an animal in an image to pass an Animal/No-Animal decision.

## 5.2. True-Negative (TN) Experiment

True-Negative (TN) images are 'No-Animal' images in the experimental set that are correctly labeled. The TN experiment investigates the rationale behind the ACS decision for the TN images. Similar to the TP experiment, we establish a hypothesis testing procedure for the assessment of the ACS decision for the TN images:

Null hypothesis ($H_0$): The ACS learns the observation sites' temporally- and statistically-dependent and independent background components, such as the background objects, shadow patterns, movement of vegetation caused by wind; Subsequently, the presence of an animal is considered as a disturbance to the learned patterns of the observation sites.

Table 7 elaborates on the algorithm for testing the null hypothesis. Again, the data preparation phase does require some attention.

*Table 7: The algorithm designed for True-Negative Experiment*

| **Experiment II: True-Negative Experimental Procedure** |
|---|
| **Input:** $N_{tn}$ experimental images correctly labeled as 'No-Animal', i.e., TN images |
| **Construct** visiting location distribution of animals in the observation sites. The center of the annotation bounding box is considered as the visiting location of the corresponding animal. |
| **Extract** three templates of two different bird species in different gestures from the TPWD images |
| **for** *TN images* $i = 1, 2, \ldots, N_{tn}$ **do** |
|     1. Introduce the first template to the $i^{th}$ TN image at a location sampled from the constructed location distribution |
|     2. Feed the new image to the ACS and collect the statistics. |
|     3. Repeat steps 1 and 2 for the second template |
|     4. Repeat steps 1 and 2 for the third template |
| **end** |
| **Using** the collected statistics: |
| **Establish** the t-test stat similar to the previous experiment |
| **Reject** the null hypothesis for a significance level $\boldsymbol{\alpha = 0.05}$ if the calculated p-value is less than $\boldsymbol{\alpha}$. |
| **Otherwise**, the experiment fails to reject the null hypothesis; this simply means that the data support the null hypothesis. A significant p-value indicates strong support for the null hypothesis. |

To introduce a disturbance in the TN images, a template of an animal is used, for which three examples are illustrated in Figure 14. Templates of two different bird species in various sitting positions are extracted from random observation sites. The bird species are chosen for imposition since birds are the smallest animals in the field of view and, thus, challenging to recognize. The animal visiting location distribution for each observation site is estimated by recording the center of the annotation bounding boxes for all animals. These distributions are demonstrated for three observation sites in Figure 15. The hypothetical birds are introduced to the TN images based on the samples from these spatial distributions.

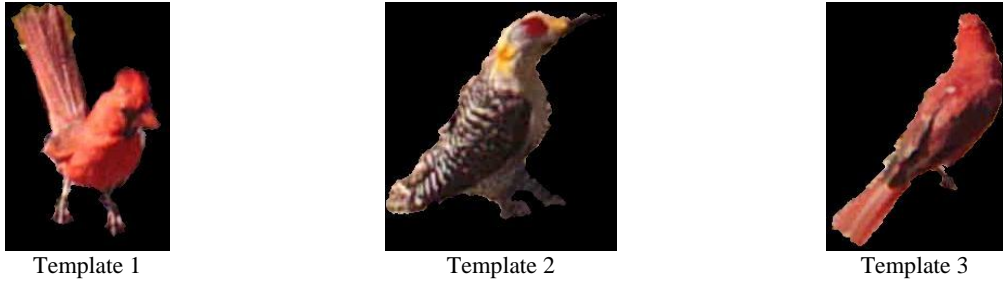

Template 1　　　　　　　　　Template 2　　　　　　　　　Template 3

*Figure 14: The three templates used in the TN experiment*

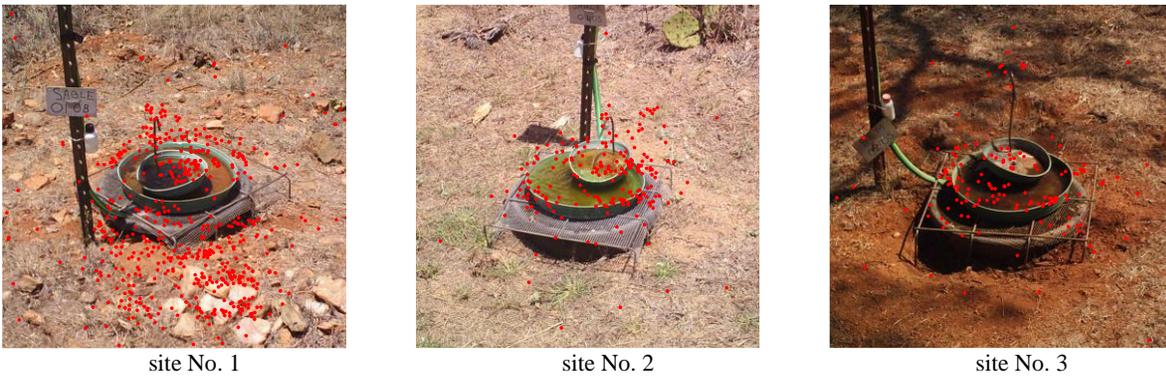

site No. 1　　　　　　　　　site No. 2　　　　　　　　　site No. 3

*Figure 15: Distribution of animal locations in the three observation sites. As shown, the animal activity is mostly concentrated around the watering fountain.*

For every TN image, the bird template is positioned in a location that is randomly sampled from the observation site's estimated animal visiting location distribution. This process is repeated for all three bird templates. Examples of disturbed TN images are shown in Figure 16.

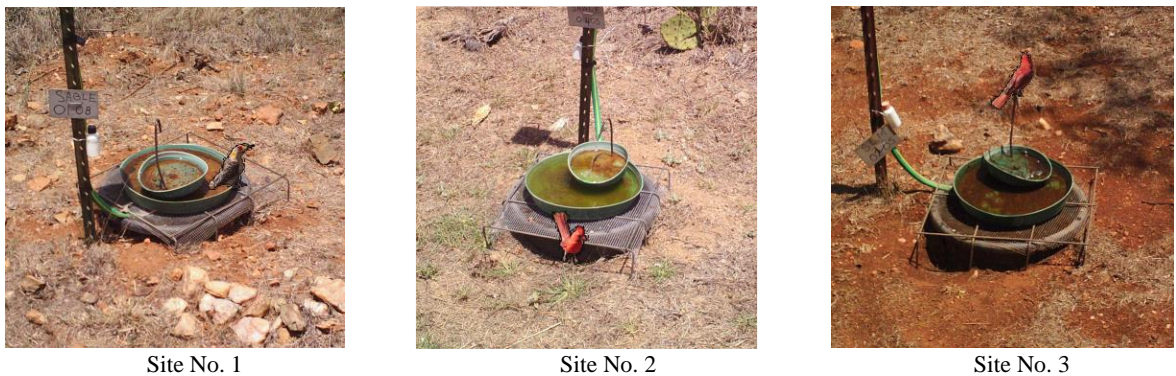

Site No. 1　　　　　　　　　Site No. 2　　　　　　　　　Site No. 3

*Figure 16: Examples of disturbed TN images in the True-Negative experiment*

The results of the ACS performance on the disturbed TN images are demonstrated in Table 8. The imposition of bird templates alters the classification label in at least 98% of the cases.

Table 8: The results of the improved True-Negative experiment. As shown, for all three observation sites, ACS estimates the "Animal" label for the disturbed TN images at least 98% of the cases.

| Obs. Site | No. TN Images | No. of Disturbed TN Images Correctly Classified | | | ACS Sensitivity on Disturbed TN Images | | |
|---|---|---|---|---|---|---|---|
| | | Template 1 | Template 2 | Template 3 | Template 1 | Template 2 | Template 3 |
| Site No. 1 | 541 | 541 | 540 | 541 | 100% | 99.8% | 100% |
| Site No. 2 | 540 | 533 | 530 | 532 | 98.7 % | 98.1% | 98.5% |
| Site No. 3 | 621 | 619 | 615 | 619 | 99.7 % | 99% | 99.7 % |
| Total | 1,702 | 1,693 | 1,685 | 1,692 | 99.5% | 99.0% | 99.4% |
| | | 1,690 | | | 99.29% | | |

Following the proposed workflow in Table 7, we test the null hypothesis:

1. The assumed success rate of the ACS for disturbed images is $p = 0.95$. The total number of samples is 1,702.
2. We model this process as a series of $N_{tn} = 1{,}702$ binomial samples, for which the expected success rate is 0.99. The samples are assumed to be independent.
3. The binomial variance, defined as $N_{tn} p(1-p) = 80.8$, is significantly larger than the threshold level of 10 [29]. This observation indicates that the binomial distribution can be approximated by a Normal distribution.

The one-sided t-test fails to reject the null hypothesis with a strong p-value of 1 and an upper bound confidence value of 0.996. Therefore, we conclude that ACS learns the background components and patterns of the observation sites and the variabilities associated with them, and the presence of an animal in fact disturbs the learned patterns of the observation sites. Thus, the classification label is determined based on whether the learned pattern is disturbed.

The p-value associated with the TN experiment is noticeably higher ($p = 1$) than the one for the TP experiment ($p = 0.63$), even though the p-value of the TP experiment is acceptable. The main reason for the notable difference of p-values is that we have a larger sample population for the TN experiment.

## 6. Bird Localization

While the ACS efficiently classifies images into 'Animal' versus 'No-Animal' categories, Birds account for more than 65% of the animal population in the TPWD database and are the most challenging to localize both manually and automatically due to their relative size, unpredictable position, and camouflage that allows them to blend in with the background (See Figure 17).

To tackle this challenge, a Bird Detection System (BDS) based on the Faster Region-based Convolutional Neural Network [30] was designed to localize the birds in the set of animal images found by the ACS. Although several published works deal with the problem of bird detection [31-38], none were found to address the aforementioned challenges in a satisfactory manner. For example, those described in [31, 36, 38] present approaches for the detection of bird parts (not birds), while others, e.g., [32, 35], focus on detection from aerial images in which the birds have significantly different radiometric and geometric appearances than those in the TPWD images.

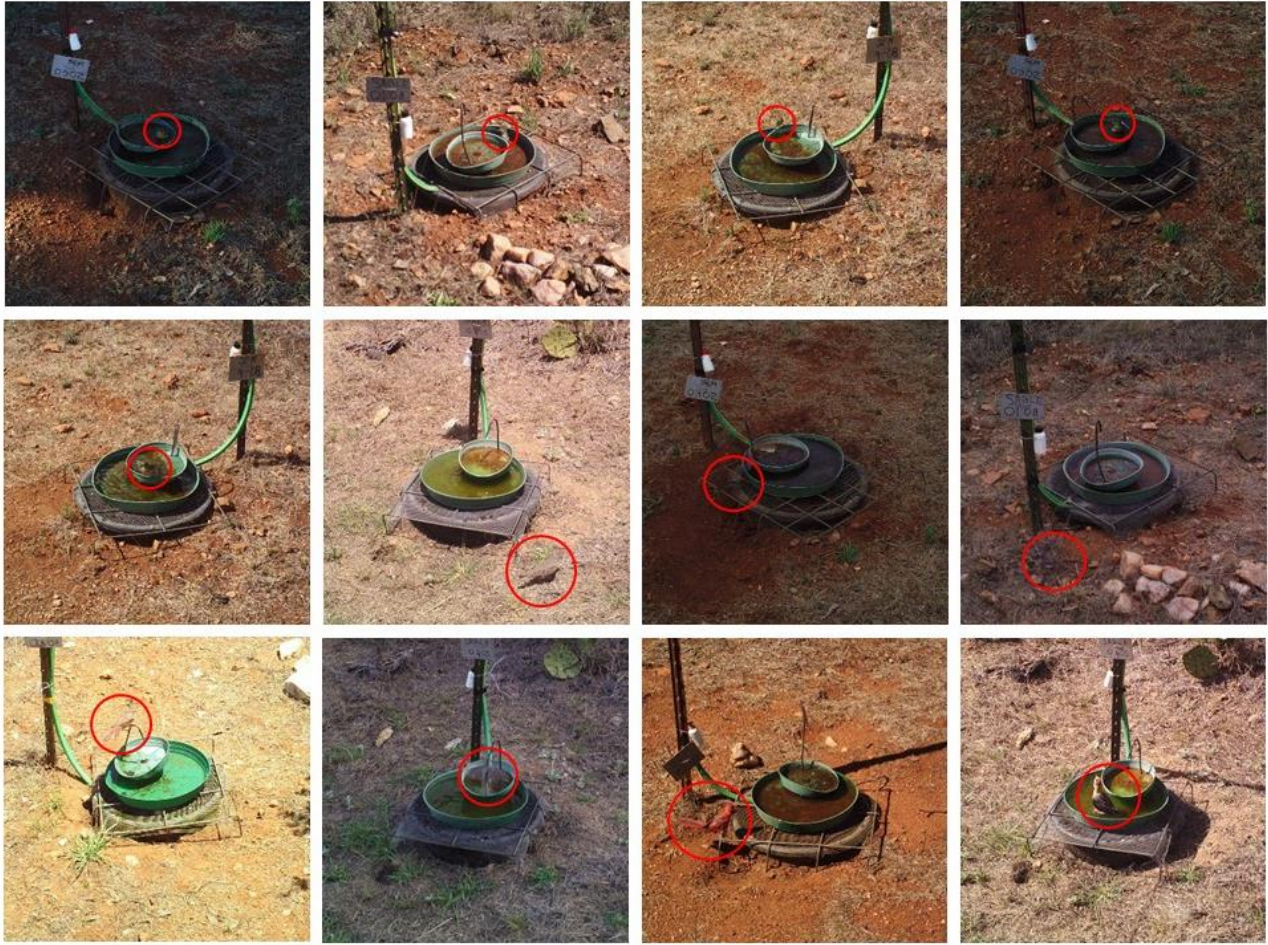

*Figure 17: Birds, which make up around 65% of the total animal population in the TPWD dataset (1,069 birds among 1,592 animals), pose a significant challenge to automatic detection because of their size (first row), camouflage (second row), position and range of activities (third row). All birds are circled in red.*

Perhaps the most relevant of existing works to that presented here are by Simons et al. [33] and Wang et al. [34]. The authors of [33] present a cascade object detector to detect and count birds in trail camera images. However, they choose not to pursue a deep learning approach and cite the small size of their training dataset as the reason. As will be detailed in the next section, our work successfully employs a deep learning strategy that was trained on a dataset even smaller than that presented in [33].

Wang et al. [34] used a modified YOLO network for bird detection trained on the 2012 PASCAL VOC dataset. A close inspection of this dataset revealed that the birds are more prominent in their respective images than the birds in our images. Specifically, while the birds occupy, on average, 18% of the image in the data used in [34], this number is as low as 1% for our images. We, therefore, concluded that the YOLO model presented in [34] could not be used to accurately localize the birds in the TPWD images.

We trained a Faster Region-based Convolutional Neural Network (Faster R-CNN) [30] to detect and localize birds in the positive images (i.e., those labelled as having animals in the ACS module). The trained network receives a preprocessed positive image and localizes the birds by estimating a bounding box per bird. In the preprocessing step, a $1500 \times 1500$ window from the input image is cropped and centered around the watering fountain. The Faster R-CNN model consists of two sub-models. The first sub-model, i.e., region proposal network, learns to find region proposals in the input image that are likely to contain a bird. Redundant RPs are eliminated by employing non-maximum suppression based on their proposal scores. The second sub-model is a classification network that ranks the selected RPs by assigning a score to every chosen RP. Finally, regions with the highest scores are outputted as bounding boxes containing birds.

To train the BDS, we generated a training set containing two subsets of images. The first subset, i.e., a positive subset, includes 80% of the single-bird images in the TPWD database. Each image is paired with a bounding box localizing the bird in the corresponding image. The second subset, i.e., negative subset, is a group of FPs collected using the hard-negative mining method [39]. Figure 18 shows examples from each of the two subsets used in training and Table 9(a) reports the details of the training set.

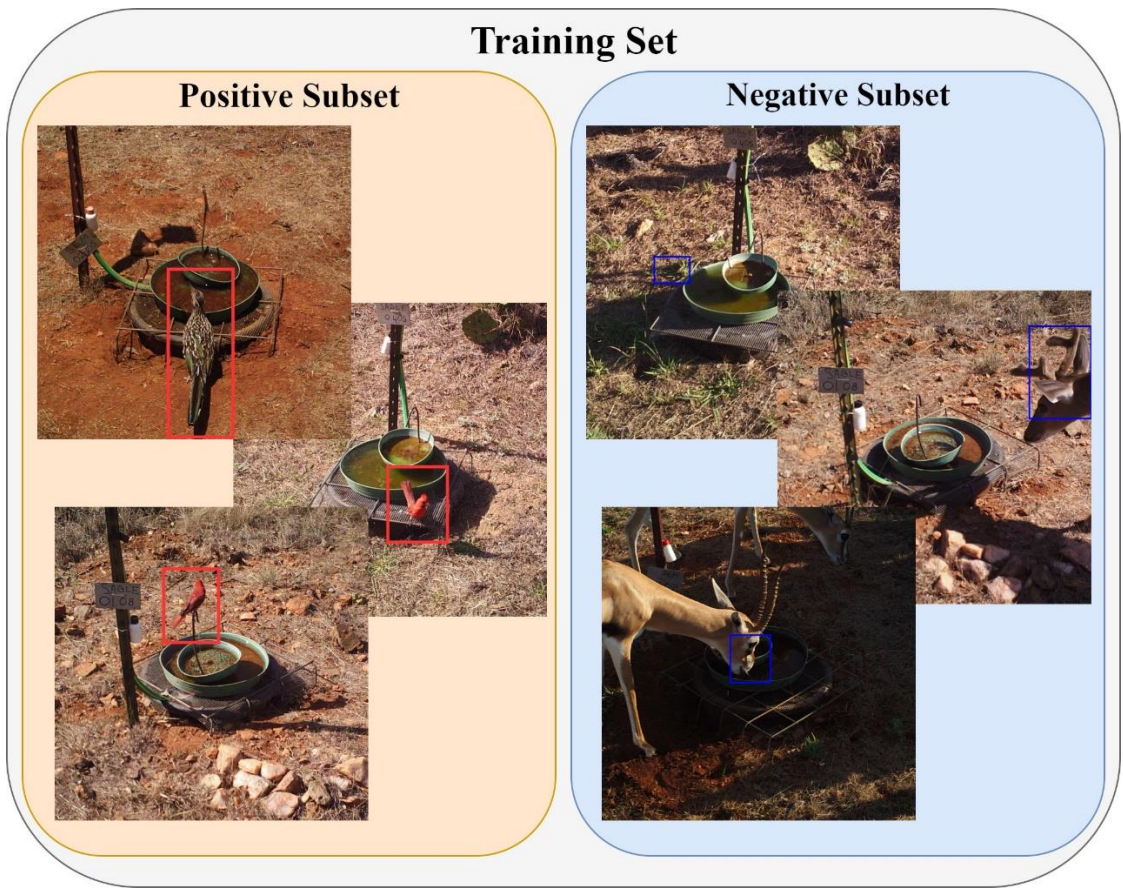

*Figure 18: Examples of images in the positive and negative subsets of the training set*

The specifications of the training procedure are shown in Table 9(c). The optimizer used is Stochastic Gradient Descent with Momentum of 0.9 and a learning rate of 0.001. For training the proposal network, a binary class label is assigned to each RP. Two kinds of RPs are assigned a positive label: (i) the RP/RPs that have the highest Intersection-over-Union (IoU) with a ground-truth box or (ii) an RP for which exists a ground-truth box with an IoU larger than 0.6. A negative label is allocated to a non-positive RP that has IoU of less than 0.5 with any ground-truth box. The training procedure of the BDS took 124 minutes on an Nvidia Titan RTX.

As illustrated in Figure 19, the trained BDS is capable of detecting single and multiple birds with varying size, color, and gesture from all observation sites.

Table 9: Details of (a) Training set, (b) Testing set and (c) Specifications of the training procedure

(a)

| Training Set Details | |
|---|---|
| No. of Single Bird Images | 855 |
| No. of No-Bird Samples | 152 |
| **Total No. of Images** | **1,007** |
| **Total No. of Birds** | **855** |

(b)

| Testing Set Details | |
|---|---|
| No. of Single-Bird Images | 244 |
| No. of Multi-Bird Images | 164 |
| No. of No-Bird Images | 1,619 |
| **Total No. of Images** | **2,027** |
| **Total No. of Birds** | **567** |

(c)

| Training Process Specifications | |
|---|---|
| Optimizer | SGDM Momentum = 0.9, Learning Rate = 0.001 |
| No. of Epochs | 10 |
| Back-bone CNN | ResNet 50 |
| Positive IoU Range | [0.6, 1] |
| Negative IoU Range | [0, 0.5] |

To quantitively assess the performance of BDS, a test set is formed that contains the remaining 20% of single bird images along with all the multi-bird images in the TPWD database. Table 9(b) tabulates the details of this test set. TPs, TNs, FPs, and FNs that are used to estimate the sensitivity and specificity of BDS are defined as follows. TPs are number of localized birds for which the IoU of the estimated and ground-truth bounding boxes is greater than 0.4. TNs are the number of no-bird images for which the model does not output an estimated localization. FPs are assessed in two ways: (1) localized birds for which the IoU of estimated and ground-truth bounding boxes is less than 0.4, and (2) number of background regions localized as a bird. Lastly, FNs are the number of birds not localized.

The BDS performed with 94% sensitivity and 93% specificity on the test set. Details are presented in Table 10.

Table 10: Details of the performance of BDS on the generated testing set

| TP | TN | FP | FN | Sensitivity | Specificity | Avg. IoU | Avg. Localization Time |
|---|---|---|---|---|---|---|---|
| 526 | 1,534 | 125 | 35 | 94% | 93% | 68% | 0.4 s (Tested on AMD Ryzen 7, 3.6 GHz) |

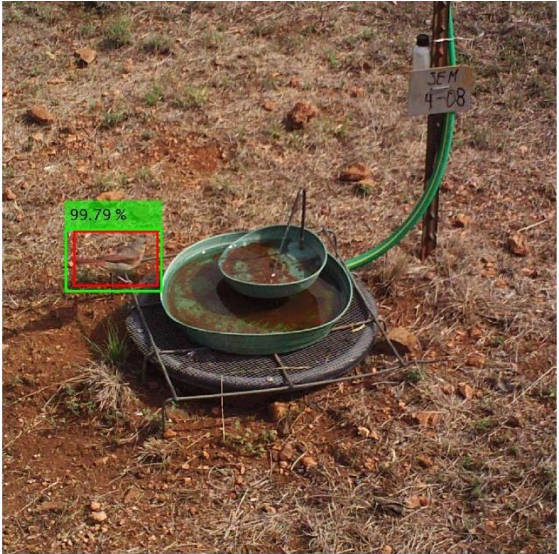
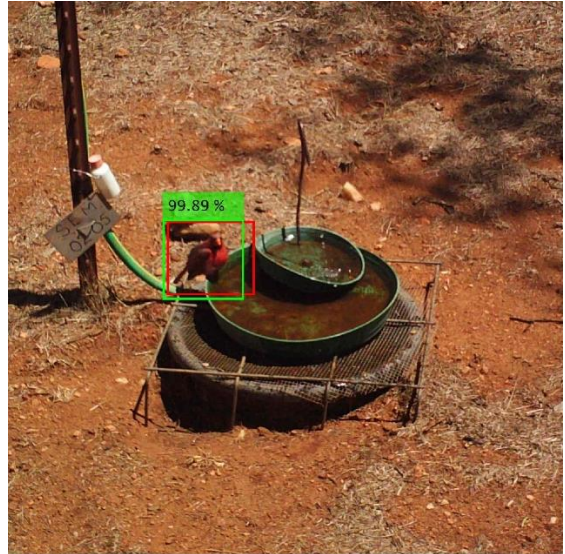
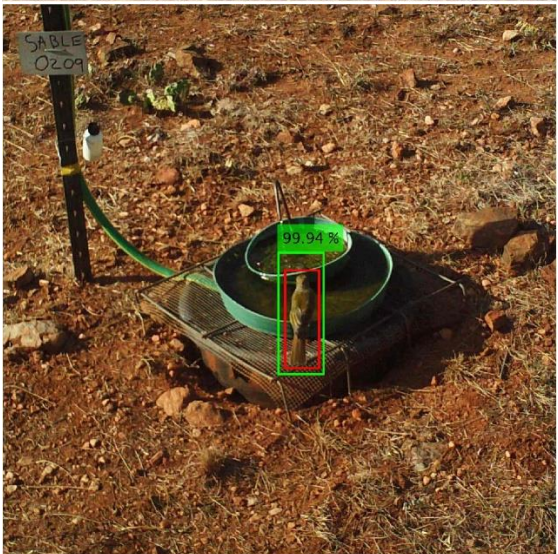
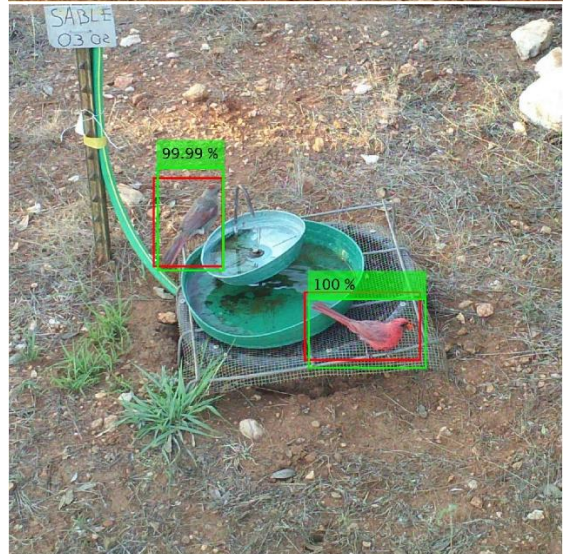
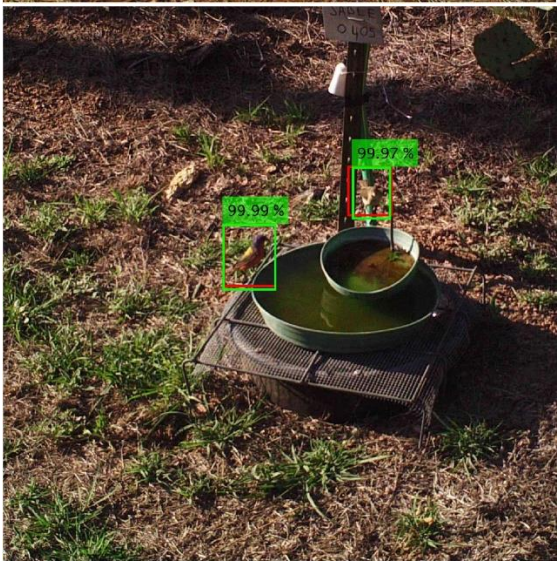
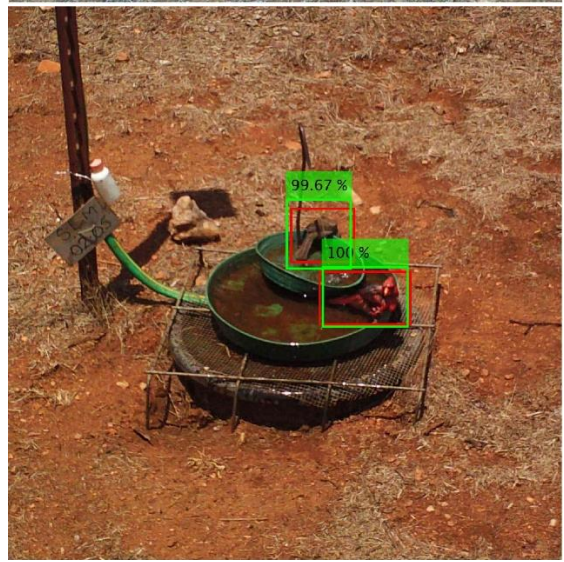

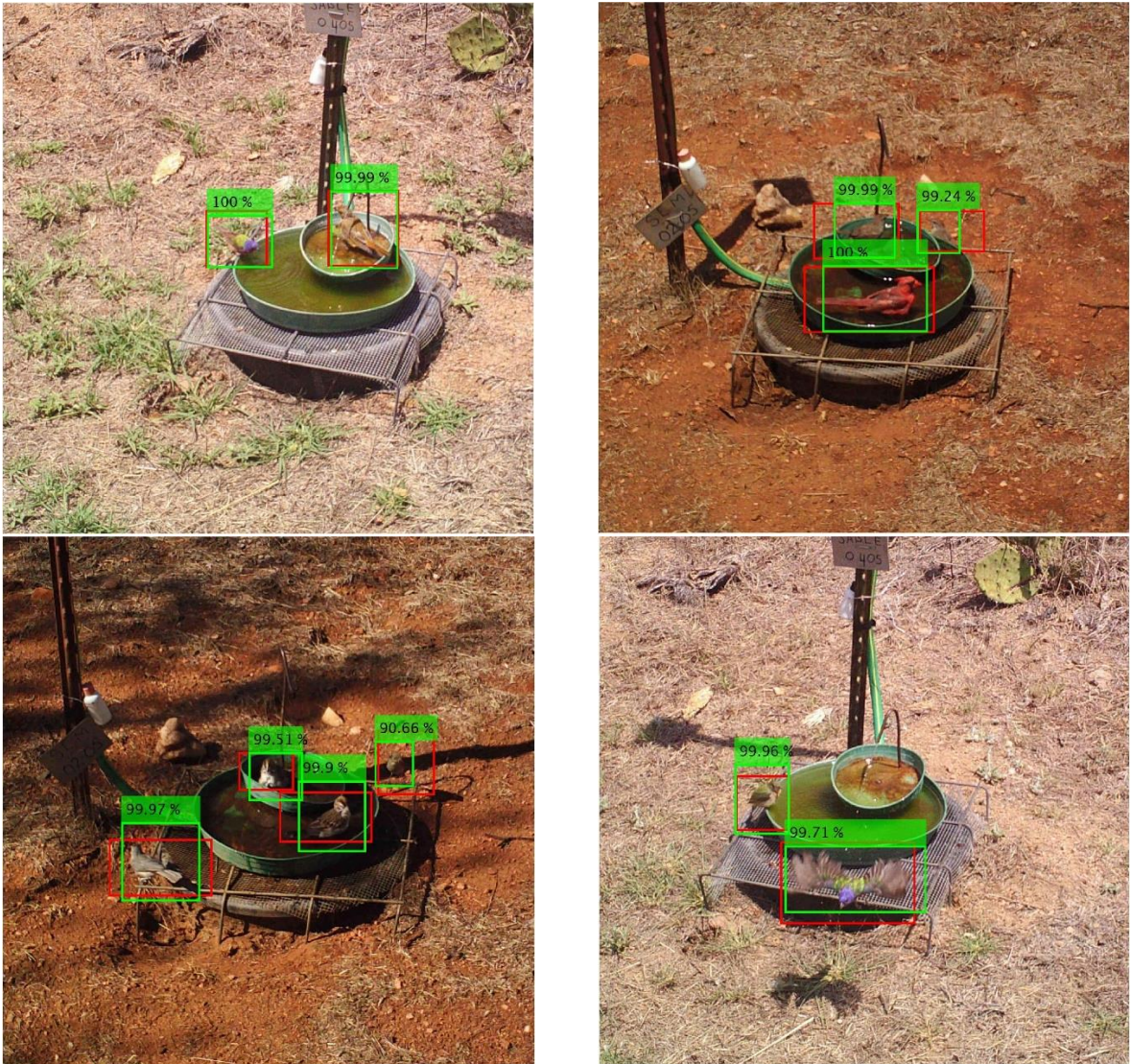

*Figure 19: Examples of birds in TPWD images localized using the BDS. Green boxes indicate the model's output, while the red boxes present the ground truth. The number in the green box headline reports the estimated detection score.*

## 7. Discussion

For our application, models from prior works in the literature performed poorly on the TPWD images ([3] [6]), indicating that DNN models that are trained to perform similar tasks may not always successfully generalize despite shared domain relevance. Mainly, we found that a carefully selected dataset was necessary to train a model to handle the variation in lighting conditions and backgrounds of observation sites in the TPWD dataset; this was evident as training on the TPWD data via random sampling proved to be insufficient for developing well-performing models.

Furthermore, seasonal and incidental changes to the scenery of the observation sites deteriorated the performance of the animal classification system over time. To maintain robust performance after deployment, it was crucial that the ACS is able to continuously recognize data drift and perform retraining when necessary.

Systems developed and deployed following the procedures outlined in this work can significantly improve and scale ecological research and conservation projects employing trail-camera imaging. On a typical 8-core CPU (AMD Ryzen 7 3700X), the classification and bird detection tasks take approximately 0.05 and 0.4 seconds per image, respectively. An image is processed through the entire pipeline in less than 0.5 seconds while a human labeler may take 30 seconds on average to accurately and consistently perform the same task. These systems accelerate otherwise costly and labor-intensive efforts by up to 60 times.

For future work, complete animal species classification may be added to the proposed pipeline; this task requires gathering more animal examples of different species to generate a multiclass dataset. Moreover, it is possible to extend the current system into one multi-stage network, e.g., a network that handles both sorting of animal vs no-animal images and the localization and species level classification of animal images. Of particular interest is the potential to employ these systems as a second opinion to verify data generated through crowdsourced labeling of trail-camera imagery, especially in cases where manual verification by domain experts isn't feasible.

## 8. Conclusion

We presented a pipeline for automatic animal classification and detection in trail-camera images taken from fields monitored by the Texas Parks and Wildlife Department. A two-stage deep learning pipeline comprising an animal classification system and a bird detection system was implemented. The animal classification system categorizes the images into 'Animal' and 'No-Animal' classes and then the 'Animal' images are processed to detect birds through the bird detection system. The animal classification system achieved an overall sensitivity and specificity of 93% and 96%, respectively. The bird detection system achieves better than 93% sensitivity and 92% specificity with an average IoU of more than 68%. These systems were shown to be useful in fast, accurate classification and detection of animals in TPWD trail-camera images. We addressed the importance of managing post-deployment data drift and updates to the CNN-based animal classification system as image features vary with seasonal changes in the wildlife habitat. For this purpose, we equipped the animal classification system with an automatic retraining algorithm that uses a novel method for inspecting drift in the incoming images and triggering the retraining process when necessary. Finally, we conducted two statistical experiments to explain the predictive behavior of the animal classification system. These experiments explored the image features that influence the system's decisions. The test results strongly supported the hypothesis that animal presence plays a critical role in the animal classification system's decision.

**Acknowledgements**

The authors would like to thank the members of the Applied Vision Lab at Texas Tech University for their assistance in image annotation, especially Peter Wharton, Rupa Vani Battula, Shawn Spicer, Farshad Bolouri, Colin Lynch, and Rishab Tewari. This research was funded by a grant from the Texas Parks and Wildlife Department.